\documentclass[runningheads]{llncs}

\usepackage{eccv}

\author{
Zejian Kang\inst{1,2} \and
Kai Zheng\inst{2} \and
Yuanchen Fei\inst{3}
Wentao Yang\inst{1,2} \and
Hongyuan Zou\inst{2,4}  \and
Xiangru Huang\inst{2}\textsuperscript{\dag}
}

\institute{
\textsuperscript{1}Zhejiang University \quad
\textsuperscript{2}Westlake University \quad
\textsuperscript{3}Hunan University \quad
\textsuperscript{4}The University of Hong Kong \\
\email{huangxiangru@westlake.edu.cn}
}

\usepackage{eccvabbrv}

\usepackage{graphicx}
\usepackage{booktabs}

\usepackage[accsupp]{axessibility}  

\usepackage{hyperref}

\usepackage{orcidlink}
\usepackage{booktabs}
\usepackage{multirow}
\usepackage[table]{xcolor}
\usepackage{algorithm}
\usepackage{algorithmic}

\usepackage{longtable}
\usepackage{booktabs}

\begin{document}

\title{SemanticFace: Semantic Facial Action Estimation via Semantic Distillation in Interpretable Space}

\titlerunning{SemanticFace}


\authorrunning{Z. Kang et al.}


\maketitle

\begin{center}
    \includegraphics[width=1\linewidth,trim=0 0 0cm 0, clip]{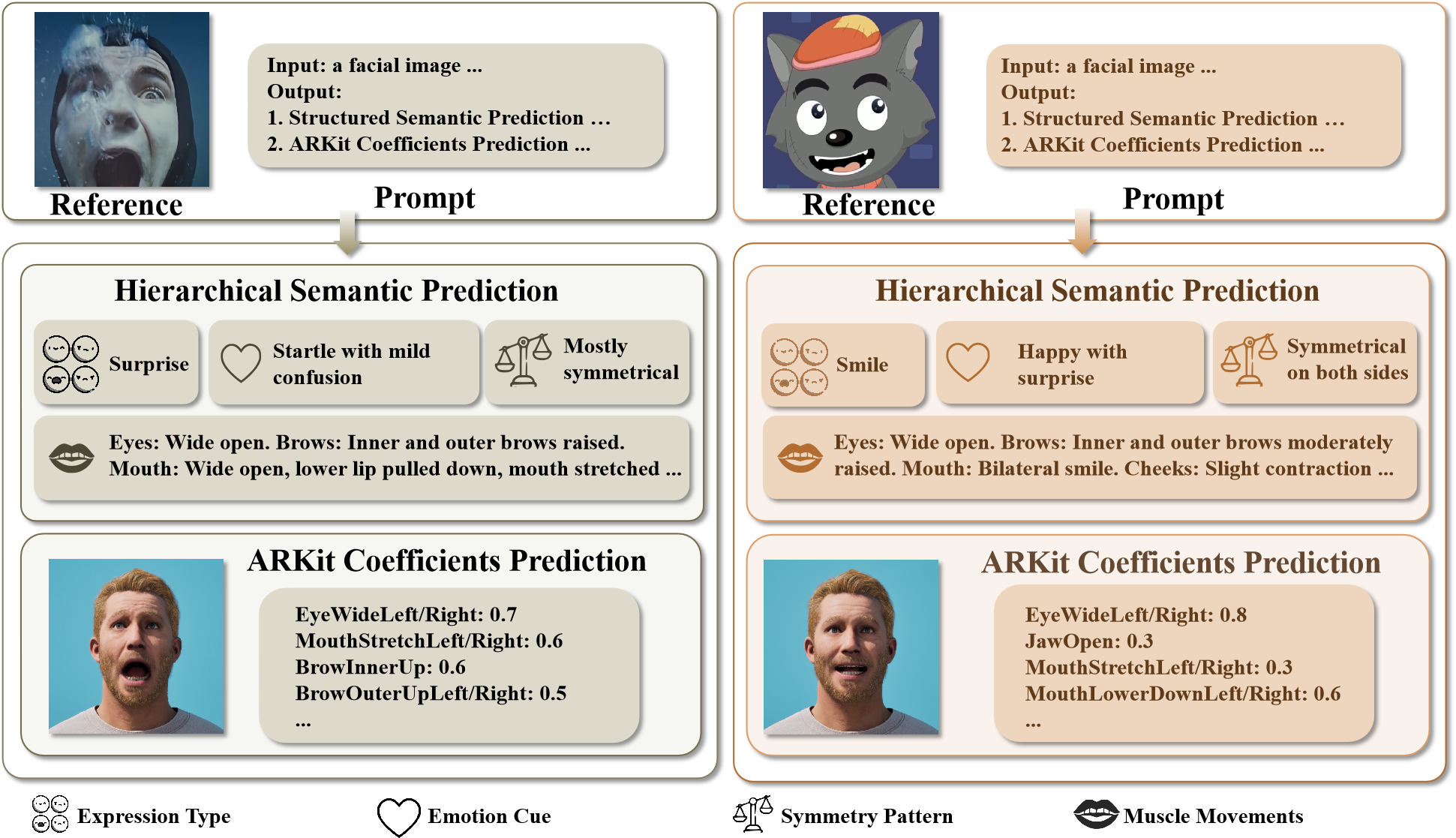}
    
    \captionof{figure}{
        \textbf{SemanticFace.}
Given a facial image, our model jointly outputs structured semantic predictions and interpretable facial action coefficients. This joint formulation bridges visual perception and semantic reasoning, yielding results that are both perceptually natural and semantically grounded, while demonstrating strong robustness across diverse identities and domain shifts (e.g., cartoon characters).}
    \label{fig:teaser}
\end{center}

\begin{abstract}


Facial action estimation from a single image is often formulated as predicting or fitting parameters in compact expression spaces, which lack explicit semantic interpretability. However, many practical applications, such as avatar control and human-computer interaction, require interpretable facial actions that correspond to meaningful muscle movements. In this work, we propose \textbf{SemanticFace}, a framework for facial action estimation in the interpretable ARKit blendshape space that reformulates coefficient prediction as structured semantic reasoning. SemanticFace adopts a two-stage semantic distillation paradigm: it first derives structured semantic supervision from ground-truth ARKit coefficients and then distills this knowledge into a multimodal large language model to predict interpretable facial action coefficients from images. 
Extensive experiments demonstrate that language-aligned semantic supervision improves both coefficient accuracy and perceptual consistency, while enabling strong cross-identity generalization and robustness to large domain shifts, including cartoon faces.  Code is available at \url{https://github.com/kangzejian1896/SemanticFace.git}.


\end{abstract}

\section{Introduction}
\label{sec:intro}

Estimating facial actions from images is a core capability for digital humans, avatar control, and human-computer interaction.
Given a single facial image, our goal is to predict interpretable blendshape coefficients that directly correspond to semantic facial movements such as jaw opening and lip corner pulling.
This problem is commonly addressed by regressing parameters in compact parametric spaces designed for geometric reconstruction and rendering consistency~\cite{danvevcek2022emoca,feng2021deca,guo2020towards,li2017flame, blanz1999morphable}.
While effective for recovering geometry, these representations are not explicitly structured as semantic facial actions, which limits their direct use as interpretable controls and makes it difficult to incorporate language-level supervision.

In many practical settings, what matters is not only geometric fidelity but also whether the predicted controls reflect the intended semantic facial actions.
Meanwhile, modern multimodal large language models (MLLMs)\cite{radford2021clip,li2023blip} encode strong visual-semantic priors, yet they require a semantically structured output interface to operate effectively.
This motivates a key question: \emph{how can we inject the visual-semantic priors of MLLMs into facial action prediction in a way that is explicit, interpretable, and trainable?}

Our key idea is to exploit an interpretable facial action space as the bridge between visual perception and language-level semantics, and to distill structured semantic knowledge into an image-to-action predictor.
Blendshape representations such as ARKit~\cite{apple_arkit,apple_livelinkface_2025},  provide such an action space, where each dimension has a predefined semantic meaning, enabling supervision beyond purely numerical targets.
Instead of treating action prediction as isolated parameter regression, we introduce \emph{semantic distillation}: converting ground-truth action coefficients into structured semantic descriptions and using them as language-aligned supervision to guide an MLLM-based predictor.

Based on this idea, we propose \textbf{SemanticFace}, a language-aligned framework for interpretable facial action estimation in the ARKit blendshape space.
SemanticFace follows a two-stage semantic distillation paradigm.
In Stage I, a pretrained large language model (LLM) converts ground-truth ARKit coefficients into hierarchical semantic descriptions that encode expression-related attributes such as muscle movement and symmetry.
In Stage II, a pretrained MLLM is adapted to predict ARKit coefficients from images under semantic supervision, using a unified autoregressive formulation. Rather than simply boosting regression accuracy with extra signals, we reinterpret facial action estimation as a language-conditioned structured generation task.

We evaluate SemanticFace on both controlled and in-the-wild settings, including comparisons with a statistical blendshape estimation method~\cite{li2026statistical} and a widely used method\cite{apple_arkit,deadface2023}.
Results show that semantic distillation improves coefficient accuracy and leads to more semantically consistent facial actions.

Our contributions can be summarized as follows:

\begin{itemize}

\item We reformulate facial action estimation as \textbf{language-aligned structured prediction} in the interpretable blendshape space, enabling semantic reasoning over facial actions.

\item We introduce a \textbf{structured semantic supervision} paradigm that converts interpretable blendshape coefficients into hierarchical semantic descriptions to expose the semantic structure of facial actions.

\item We propose a \textbf{language-prior semantic distillation} framework that transfers semantic knowledge from a pretrained LLM to a multimodal image-to-action predictor.

\item Extensive experiments demonstrate improved coefficient accuracy and perceptual consistency, and show strong cross-identity generalization even when trained on a limited number of subjects, including challenging \textbf{in-the-wild} scenarios and stylized domains such as cartoon characters.

\end{itemize}  
\section{Related Work}

\subsection{Face Modeling} 
Facial expression modeling has shifted from computationally expensive high-resolution meshes to compact parametric representations that balance efficiency and expressiveness.
Early 3D Morphable Models (3DMMs)~\cite{blanz1999morphable} employed principal component analysis (PCA) to capture common facial features via linear combinations, reducing complexity while enabling realistic face generation. 
Blendshapes extended this paradigm to dynamic expressions, offering a semantically interpretable parameterization where parameters correspond to localized muscle actions~\cite{park1974parametric,lewis2014practice}. While FLAME\cite{li2017flame} offers high compactness and has been widely adopted, its PCA-based expression space suffers from limited semantic disentanglement. Its expression parameters, derived from PCA, show poor semantic separation\cite{josi2025serep}, which limits compatibility with language-aligned supervision from MLLM~\cite{radford2021clip,liu2023llava}. In contrast, frameworks such as ARKit~\cite{apple_arkit} and MHR~\cite{ferguson2025mhr} preserve explicit semantic features, aligning well with language-based reasoning in the MLLM era. ARKit, in particular, facilitates easy data acquisition via smartphone cameras with depth sensors\cite{menzel2022automated,Aloni_2025_ICCV,wu2025keyframefacetextexpressivefacial}, unlike the more labor-intensive MHR. Although ARKit-specific advancements lag behind FLAME, insights from the latter can inform transferable strategies for semantic-aware estimation.

\subsection{Facial Action Estimation} 
Single-image parametric estimation has progressed from sparse landmark-based methods to deep learning approaches and recently to leveraging large pretrained model.
Early techniques primarily relied on facial landmarks, aligning sparse or dense keypoints with template meshes via statistical methods such as Active Shape Models (ASM)~\cite{cootes1995active}, Active Appearance Models (AAM)~\cite{cootes2001active,edwards1998interpreting}, or 3D Morphable Model fitting~\cite{blanz2003face,matthews2004aam}, as well as early single-image 3D reconstruction techniques~\cite{kemelmacher2011face,hassner2013viewing,zhu2016face}. More recent landmark detectors like BlazeFace~\cite{bazarevsky2019blazeface} and attention-based approaches~\cite{koetzier2020attention} have also been employed. However, their sparsity limits robustness to extreme poses or expressions~\cite{taubner2024flowface}. 
Subsequent works shifted to convolutional neural networks (CNNs), typically using ResNet-50 encoders to extract high-dimensional features followed by regression heads~\cite{he2016resnet,howard2017mobilenets}, incorporating refinements for emotional details~\cite{danvevcek2022emoca,feng2021deca} and differentiable rendering for self-supervision~\cite{retsinas2024smirk,sanyal2019ringnet}. 
Despite improvements, CNNs often overlook global context, and the small parameter scale constrains detail capture and generalization. Recent data-driven methods leverage large pretrained models to inject stronger priors: for instance, Vision Transformers (ViTs) generate spatial-geometric constraints~\cite{dosovitskiy2021vit,giebenhain2025pixel3dmm}, while alternatives like LRM extract features as ResNet substitutes~\cite{zhu2026kaolrm}, and LAM integrates Gaussian attributes for enhanced expressivity~\cite{he2025lam}. This trend underscores a shift toward exploiting vast pretraining data for more comprehensive feature extraction. ARKit-based parametric estimation remains underdeveloped but follows similar technical trajectories as FLAME, yet underutilizes its strong semantic properties.
Existing ARKit works include direct CNN regression as in DNPM~\cite{cao2024dnpm}, geometry-driven landmark alignment~\cite{li2026statistical}, and early attempts at incorporating large pretrained models for feature encoding or generation~\cite{zhang2023dreamtalkdiffusionbasedrealisticemotional}.
However, most approaches largely ignore ARKit’s key advantage-its coefficients are already designed to be human-interpretable actions~\cite{wolff2024mapping} (e.g., \textit{
BrowInnerUp}, \textit{EyeBlinkLeft}), which opens a natural path for MLLMs to perform structured, semantically grounded reasoning instead of black-box regression. 

\section{Method}
\label{sec:method}

\begin{figure}[t] 
    \centering
    \includegraphics[width=0.95\linewidth]{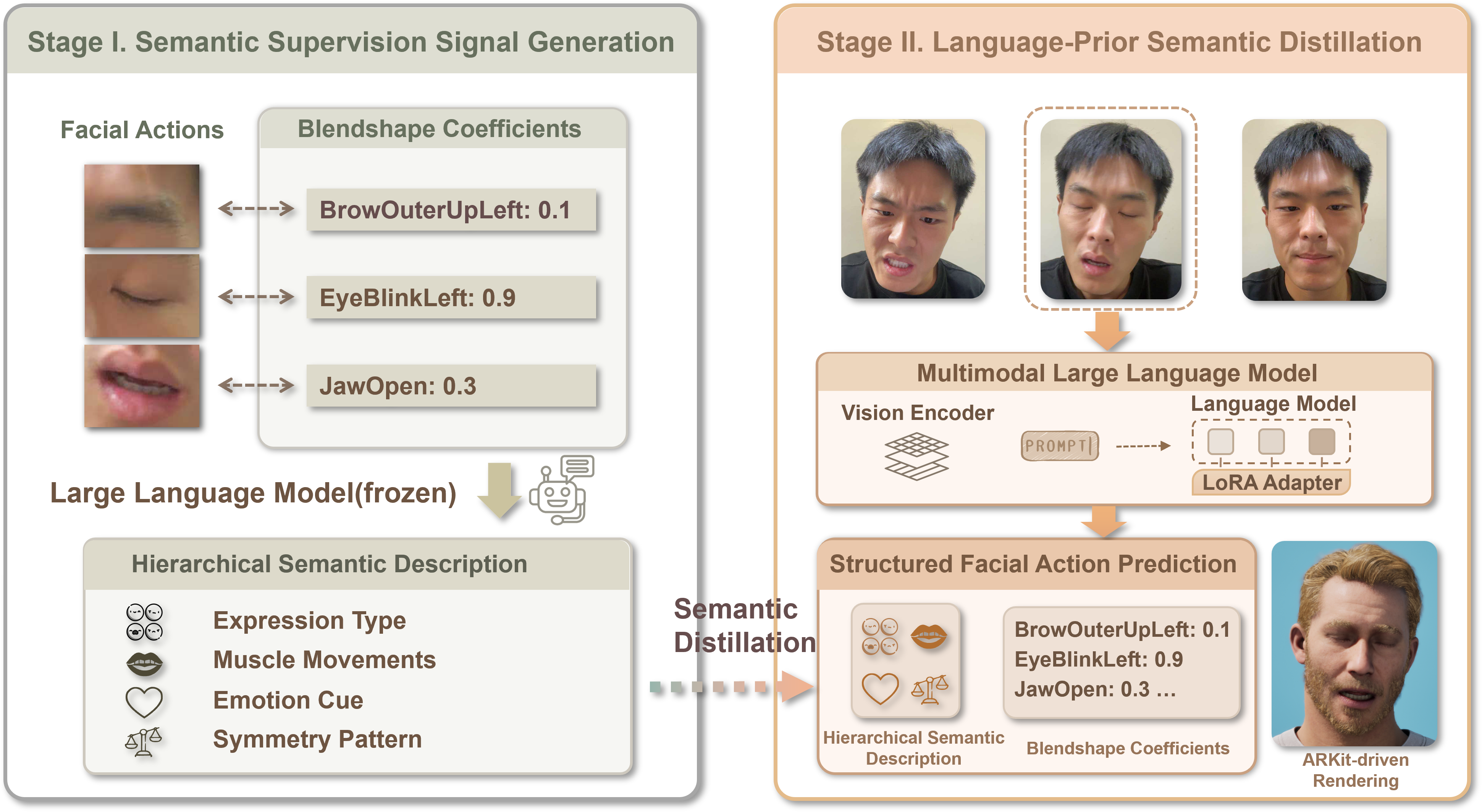} 
    \caption{
\textbf{Overview of the SemanticFace framework.}
In Stage I, a frozen LLM serves as a \emph{semantic teacher}, converting ground-truth ARKit blendshape coefficients into hierarchical semantic descriptions.
In Stage II, these structured language-aligned representations are distilled into a MLLM acting as the \emph{student}.
Through language-prior semantic distillation, the student learns to predict ARKit coefficients from images under structured semantic guidance within an interpretable facial action space.}
    \label{fig:semanticFace}
\end{figure}


Our task is to predict ARKit blendshape coefficients from a single facial image, as formally defined in \cref{sec:problem_formulation}. 
SemanticFace follows a two-stage semantic distillation paradigm. 
In Stage I, a pretrained LLM serves as a \emph{semantic teacher}, converting ground-truth ARKit coefficients into structured hierarchical semantic descriptions (\cref{sec:stage1}). 
In Stage II, a pretrained MLLM acts as the \emph{student}, and is adapted to generate ARKit coefficients from input images under the guidance of the teacher-provided semantic supervision (\cref{sec:stage2}). 
An overview is shown in \cref{alg:semantic_distillation} and  \cref{fig:semanticFace}.






\subsection{Problem Formulation}

\label{sec:problem_formulation}

Given an input image \( I \in \mathbb{R}^{H \times W \times 3} \),
our target is to predict ARKit blendshape coefficients
\( \mathcal{V} = \{v_k\}_{k=1}^{K} \) with \( K=61 \).
Each coefficient \( v_k \in \mathbb{R} \) corresponds to the value of a predefined facial action
\( a_k \in \mathcal{A} \),
where \( \mathcal{A} = \{a_k\}_{k=1}^{K} \) denotes the ARKit blendshape action set.
We represent the prediction as a structured action-value set
\( \mathcal{S} = \{(a_k, v_k)\}_{k=1}^{K} \),
which explicitly associates each facial action with its activation level.

Unlike latent expression spaces,
ARKit defines an interpretable action space with explicit semantic meaning.
This structure enables language-aligned supervision,
which forms the foundation of our method.

\begin{algorithm}[h]
\caption{Training and inference of the proposed SemanticFace framework}
\label{alg:semantic_distillation}
\begin{algorithmic}[1]

\REQUIRE Training images $\{I_i\}_{i=1}^{N}$, ARKit action set $\mathcal{A}=\{a_k\}_{k=1}^{K}$,
ground-truth value sets $\{\mathcal{V}^{gt}_i\}_{i=1}^{N}$, pretrained LLM, pretrained MLLM
\ENSURE Trained image-to-ARKit generator

\STATE \textbf{Stage I: Semantic Supervision Signal Generation}
\FOR{each training sample $(I_i, \mathcal{V}^{gt}_i)$ with action set $\mathcal{A}$}
    \STATE Construct ground-truth action-value set:
    \STATE \quad $\mathcal{S}^{gt}_i \leftarrow \{(a_k, v^{gt}_{i,k}) \mid a_k \in \mathcal{A}\}$
    \STATE Generate semantic description:
    \STATE \quad $T^{gt}_i \leftarrow \mathrm{LLM}(\mathcal{S}^{gt}_i)$
\ENDFOR

\STATE

\STATE \textbf{Stage II: Language-Prior Semantic Distillation}
\FOR{each training sample $(I_i, \mathcal{S}^{gt}_i, T^{gt}_i)$}
    \STATE Construct training target:
    \STATE \quad $U^{gt}_i \leftarrow [T^{gt}_i \; || \; \mathcal{S}^{gt}_i]$
    \STATE Update MLLM parameters using autoregressive loss:
    \STATE \quad $\mathcal{L} = - \sum_t \log p_\theta(u_t \mid u_{<t}, I_i)$
\ENDFOR

\STATE

\STATE \textbf{Inference}
\FOR{each test image $I$}
    \STATE Predict description and coefficient sequence using trained MLLM:
    \STATE \quad $\hat{U} \leftarrow \mathrm{MLLM}(I)$
    \STATE Parse predicted value set $\hat{\mathcal{V}}$ from $\hat{U}$
\ENDFOR

\end{algorithmic}
\end{algorithm}

\subsection{Semantic Supervision Signal Generation}
\label{sec:stage1}

Facial actions exhibit structured semantic relationships,
including regional coordination, symmetry patterns,
and global expression configuration. To capture such structure, we generate
structured semantic supervision from ground-truth ARKit coefficients
using a pretrained LLM.

Formally, given the ARkit blendshape coefficients 
$\mathcal{V}^{gt}=\{v_k^{gt}\}_{k=1}^{K}$ associated with the action set
$\mathcal{A}=\{a_k\}_{k=1}^{K}$,
we represent the coefficients as the structured action-value set
$\mathcal{S}^{gt} = \{(a_k, v_k^{gt})\}_{k=1}^{K}$.
A structured semantic description is then generated as
\begin{equation}
T^{gt} = \mathrm{LLM}(\mathcal{S}^{gt}),
\end{equation}
where $T^{gt}$ encodes hierarchical semantic attributes
(e.g., expression category, coordinated muscle activations,
and symmetry patterns) that are implicit in the numerical parameter space.

This process generates an intermediate semantic representation that bridges the low-level parameter space and high-level facial reasoning.
Rather than supervising only scalar targets,
we supervise the model with structured semantic knowledge distilled from coefficients. The structured semantic description $T^{gt}$ is hierarchical
and contains the following components:

    
    
    

\begin{itemize}
    \item \textbf{Expression Type:}
    structural configuration of facial actions (e.g., \textit{smiling}).

    \item \textbf{Muscle Movements:}
    localized action activations corresponding to specific muscle groups.

    \item \textbf{Emotion Cue:}
    affective interpretation only when clearly supported by observable actions (e.g., \textit{happy}).

    \item \textbf{Symmetry Pattern:}
    spatial coordination patterns across bilateral actions.
\end{itemize}

Unlike scalar parameter targets,
$T^{gt}$ explicitly encodes relationships among actions
at multiple semantic levels.
It captures how regional movements compose global expressions,
and how bilateral actions interact.


Although the semantic descriptions are deterministically derived from ground-truth coefficients, they impose an explicit structural prior over action relationships that is otherwise absent in independent scalar supervision. The role of the LLM is not to provide additional labels, but to externalize latent structural dependencies among facial actions into a language-aligned form. These structured semantic representations are then distilled into an image-conditioned model to guide interpretable facial action prediction.

\subsection{Language-Prior Semantic Distillation}
\label{sec:stage2}

We treat the pretrained LLM as a semantic teacher
that encodes structured relationships among facial actions,
and the MLLM as a student
that learns to predict interpretable facial actions from images.
We adopt a pretrained MLLM consisting of a vision encoder
and a causal language decoder.
Given an image $I$, the vision encoder produces visual tokens,
which condition the decoder to generate structured outputs.

To distill semantic knowledge from the teacher,
we supervise the student to jointly generate
(i) the structured semantic description $T^{gt}$ produced by the LLM,
and (ii) the action–value sequence $\mathcal{S}^{gt}$.
The combined target sequence is
\begin{equation}
U^{gt} = [T^{gt} \; || \; \mathcal{S}^{gt}].
\end{equation}

We implicitly transfer structured semantic relationships from the LLM to the image-conditioned predictor by requiring the MLLM to reproduce both semantic descriptions and action-value sequences.
This process injects language level priors
about regional coordination, symmetry, and expression configuration
into the parameters of the MLLM.


Unlike auxiliary multi-task supervision, the semantic sequence is not treated as an independent prediction target. Instead, it restructures the output space into a causally ordered, semantically constrained generation process. This alters the inductive bias of the predictor from independent scalar estimation to structured action reasoning.

\section{Experiments}
\label{sec:experiments}

We conduct comprehensive experiments to systematically evaluate the proposed framework. We first describe the data collection protocol (\cref{sec:dataset}). We then present the experimental settings (\cref{sec:settings}), followed by implementation details (\cref{sec:Implementation_details}). Next, we report quantitative and qualitative results on the test set (\cref{sec:test_set}). To further assess robustness and generalization, we evaluate performance under in-the-wild conditions (\cref{sec:in-the-wild}). We also provide qualitative comparisons with FLAME-based methods (\cref{sec:flame}). Finally, we conduct ablation studies to quantify the contribution of hierarchical semantic supervision (\cref{sec:ablation}).

\subsection{Dataset}
\label{sec:dataset}


\subsubsection{Data Acquisition.}
We utilized an iPhone with a depth camera and the Live Link Face system \cite{apple_livelinkface_2025} to capture ARKit expression coefficients with frame-level accuracy at a consistent rate of 60 frames per second.
To ensure the purity of the expression data and eliminate individual physiological variances, we implemented a rigorous calibration protocol. 
For every recording session, a ``neutral face'' was first captured to serve as a topological and expressive baseline. 
All subsequent raw expression sequences were calibrated against this baseline. 
This normalization process ensures that the resulting coefficients represent pure expressive motion, independent of the subject’s resting facial structure. 

\subsubsection{Dataset Statistics and Diversity.}




The dataset is characterized by its scale, linguistic diversity, and professional quality performances.
It contains recordings from nine professional actors, each performing 100 scripted sequences, resulting in a total of 900 sequences and approximately 11.5 hours of high resolution video data.
The dialogue scripts were generated with GPT-5~\cite{openai_gpt5_2025}, enabling broad coverage of diverse emotional expressions. 
Each sample contains synchronized video, audio, and ARKit coefficients.
Specifically, the dataset provides high-definition facial video, clean vocal tracks aligned with facial movements, and calibrated 61-dimensional ARKit coefficients, including 52 blendshape coefficients and 9 head-motion coefficients. We therefore prioritize the diversity of facial expressions and speech-driven behaviors rather than identity variation.

Furthermore, we analyze the emotional diversity of the dataset to verify that it covers a wide range of expressive states. We employ the LLM Qwen1.5‑1.8B\cite{qwen1.5} to infer the emotional distribution of each script line. Given the script title and dialogue, the model outputs a seven-dimensional probability vector representing the intensity of seven basic emotions (happiness, neutral, sadness, anger, fear, surprise, and disgust). The resulting vector is normalized to ensure the sum equals one and is used as the emotional representation of the script. The dominant emotion of each sequence is determined by the highest intensity dimension.

\begin{figure}
    \centering
    \includegraphics[width=0.9\linewidth]{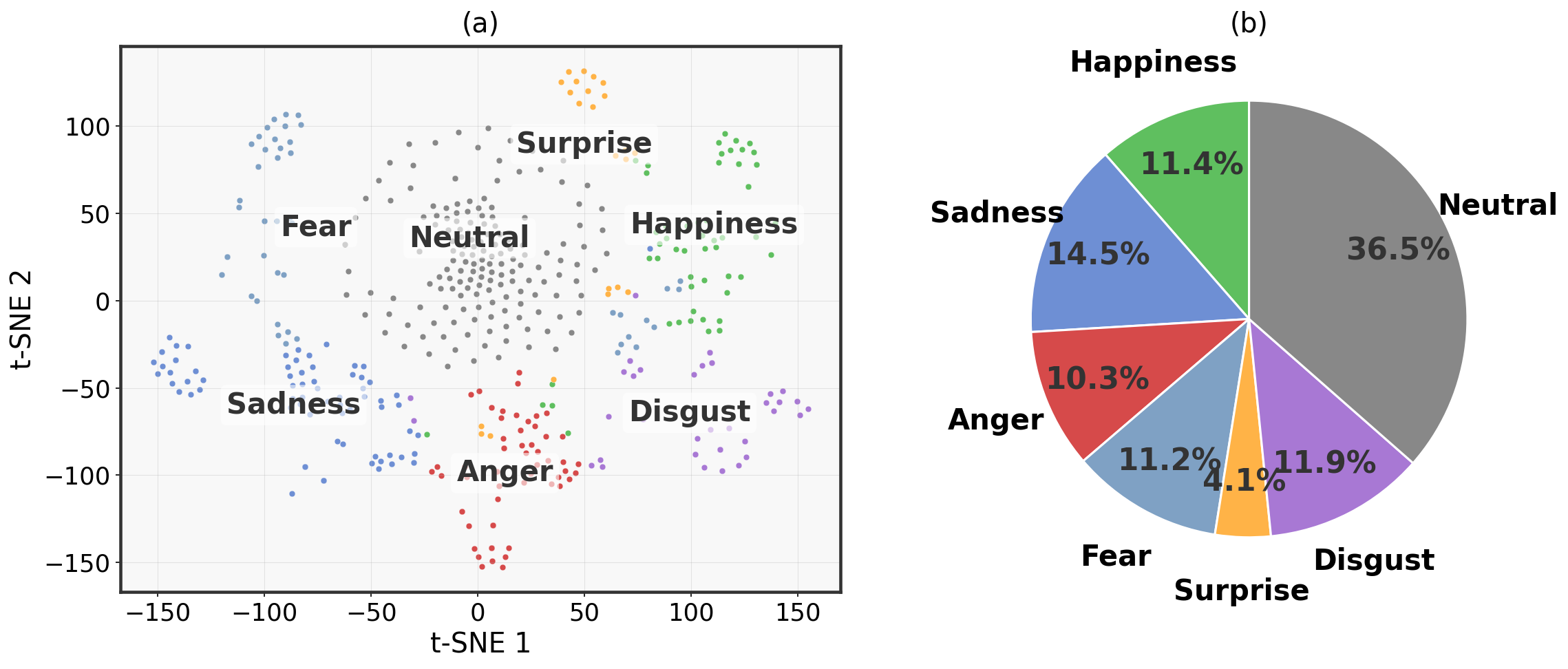}
    \caption{\textbf{Emotion Distribution and Visualization.} (a) t-SNE projection of scripts
colored by emotion intensities. (b) Distribution of dominant emotions across all
scripts.}
    \label{fig:emotion}
\end{figure}

As illustrated in \cref{fig:emotion}, we project the emotion vectors into a two-dimensional space using t-SNE. Distinct clusters corresponding to different emotional categories emerge clearly, indicating that the dataset contains semantically separable emotional expressions. Neutral samples occupy the central region, while more expressive emotions such as anger, sadness, and happiness form well-separated clusters. The distribution of dominant emotions further shows that the dataset maintains a balanced coverage of emotional categories, demonstrating that it spans a broad spectrum of affective states rather than concentrating around a limited subset of expressions. Such diversity exposes the model to a wide variety of facial expression dynamics during training.

 \subsubsection{Data Processing.}
Our dataset consists of facial videos paired with ARKit blendshape coefficients.
To reduce temporal redundancy, we subsample the original videos at 2 frames per second, as adjacent frames often exhibit highly similar facial configurations and provide limited additional supervisory signals.
This sampling strategy improves data efficiency while preserving diverse expression states. 
We adopt a subject-disjoint split, using 7 subjects for training (700 sequences, 28,209 synchronized image-ARKit coefficient pairs), 1 subject for validation (100 sequences, 4,960 synchronized pairs), and 1 subject for testing (100 sequences, 4,700 synchronized pairs).
Each subject performs a diverse set of scripted expressions covering multiple emotion categories, resulting in substantial intra-subject variation.
This procedure ensures that identities in the test set are unseen during training, enabling rigorous evaluation of cross-identity generalization.

\subsection{Experimental Setting}
\label{sec:settings}

\subsubsection{Evaluation Metrics.}
\label{appendix:evaluation_metrics}


We employ several metrics to assess performance across different dimensions:
Mean Squared Error (MSE), R-precision, Multimodal Distance (MMD), and Cross-Comparison.
MSE measures the average squared difference between predicted blendshape coefficients and the ground-truth ARKit coefficients.
R-Precision evaluates retrieval performance by checking whether the ground-truth motion corresponding to a given image description appears among the top-$K$ ($K=3$) ranked candidates within a batch of 32 samples~\cite{Aloni_2025_ICCV,wu2025keyframefacetextexpressivefacial}.
MMD measures the distributional discrepancy between image and motion embeddings in the learned joint embedding space.
Cross-Comparison, following the protocol introduced in prior work~\cite{li2026statistical}, provides a benchmark comparing predicted blendshape coefficients against ARKit ground truth~\cite{apple_arkit}.
The evaluation focuses on the 13 most dominant and expressive blendshapes, which serve as primary indicators of facial expression quality.
As the implementation of~\cite{li2026statistical} is not publicly available, we report the values provided in their paper and use them as a reference for comparison.
Detailed explanations of the metrics are provided in the Appendix.



\subsubsection{Baselines.}

We compare SemanticFace with two representative baselines for ARKit blendshape estimation.
SBCA~\cite{li2026statistical} is a statistical approach that regresses blendshape coefficients from facial landmarks using affine transformations, segmentation, and regression models.
DeadFace~\cite{deadface2023} is an open-source implementation built upon MediaPipe~\cite{lugaresi2019mediapipe,kartynnik2021realtime}, providing facial geometry estimation from monocular images.
Both methods represent geometric approaches that directly regress ARKit coefficients from images without leveraging semantic priors from MLLM.

We evaluate all methods on our test set with paired 3D ground-truth ARKit coefficients in \cref{sec:test_set}. 
For the in-the-wild setting without ground-truth annotations (\cref{sec:in-the-wild}), we sample 660 images from the “9 Facial Expressions for YOLO” dataset~\cite{nazmus_sakib_2025} as test data. 
Since no 3D ground-truth ARKit coefficients are available, we assess semantic alignment using R-Precision and distributional consistency using MMD.

\subsection{Implementation details}
\label{sec:Implementation_details}

Our framework employs different pretrained foundation models for the two stages.
In Stage I (semantic supervision signal generation), we use the LLM Qwen3-14B~\cite{qwen3} to convert ground-truth ARKit coefficients into structured semantic descriptions.
The LLM is kept frozen during this stage and used solely for semantic supervision generation, ensuring stable and consistent semantic annotations.
Semantic descriptions are generated offline prior to Stage II training and reused across epochs without additional computational overhead.

In Stage II (language-prior semantic distillation), we build upon the pretrained MLLM Qwen3-VL-4B-Instruct~\cite{qwen3vl} and adapt it using parameter-efficient updates via Low-Rank Adaptation (LoRA)~\cite{hu2021lora}.
Specifically, LoRA modules are applied to the language decoder, while the vision encoder and multimodal alignment modules are kept frozen to preserve pretrained multimodal representations and maintain training stability.
This design enables efficient adaptation while retaining strong multimodal priors from large-scale pretraining.

The model is trained for 50 epochs using the AdamW~\cite{loshchilov2017decoupled} optimizer with a learning rate of $1\times10^{-4}$.
Training is conducted on 8 NVIDIA A100 GPUs, with an effective batch size of 4 samples per GPU.
The total training time is approximately 45 hours.

During training, the model jointly generates distilled semantic descriptions and ARKit action–value sequences under a unified autoregressive objective, enabling alignment between visual perception and the interpretable facial action space.
For visualization, predicted ARKit coefficients are rendered using MetaHuman~\cite{metahuman}, a high-fidelity digital human system that enables realistic facial animation from blendshape coefficients.

\subsection{Evaluation on the Test Set}
\label{sec:test_set}

\begin{table}[h]
\centering
\caption{\textbf{Quantitative comparison with different methods on the test set.} 
Best values (excluding GT-test) are highlighted in \colorbox{red!20}{red}. 
The GT-test row is shown with a gray background for reference and is not included in the comparison.}
\label{tab:sota_comparison}
\scriptsize
\setlength{\tabcolsep}{0pt}  
\begin{tabular}{l c ccc c ccccc}
\toprule
\multirow{2}{*}{\textbf{Method}}
& \multicolumn{1}{c}{\textbf{Error}}
& \multicolumn{3}{c}{\textbf{R-Precision} $\uparrow$}
& \multirow{2}{*}{\textbf{MMD} $\downarrow$}
& \multicolumn{5}{c}{\textbf{Main Cross-Comparison}} \\
\cmidrule(lr){2-2} \cmidrule(lr){3-5} \cmidrule(lr){7-11}
& \textbf{MSE} $\downarrow$ 
& \textbf{top-1} & \textbf{top-2} & \textbf{top-3} 
& 
& \textbf{P-Cor} & \textbf{S-Cor} & \textbf{Acc.} & \textbf{MSD} & \textbf{Deviation} \\
\midrule
\rowcolor{gray!15}
GT-test
& 0 
& 24.28\% & 38.60\% & 48.48\% 
& 1.00 
& 100\% & 100\% & 1 & 0 & 0 \\
\midrule
DeadFace\cite{deadface2023}
& 0.0316 
& 13.43\% & 23.06\% & 30.25\% 
& 2.12 
&68.48\%&\cellcolor{red!20} 67.03\%&0.9836&0.0094&0.0691\\
SBCA\cite{li2026statistical}
& 0.0190 
& - & - & - 
& - 
& 67.44\% & 65.61\% & 0.8449 & 0.0298 & 0.1470  \\
\midrule
Ours
& \cellcolor{red!20} 0.0034
& \cellcolor{red!20} 22.15\%
& \cellcolor{red!20} 37.06\%
& \cellcolor{red!20} 48.06\%
& \cellcolor{red!20} 1.18 
& \cellcolor{red!20} 70.50\%
& 63.32\%
& \cellcolor{red!20} 0.9967 
& \cellcolor{red!20} 0.0060
& \cellcolor{red!20} 0.0554 \\
\bottomrule
\end{tabular}
\end{table}



Quantitative results on coefficient prediction accuracy are reported in \cref{tab:sota_comparison}. 
SemanticFace consistently outperforms existing baselines across all comparable metrics beside S-Cor.
In a few ambiguous cases, semantic distillation may slightly adjust the relative strength of similar actions to better match the overall semantic meaning (e.g., expression category). This small re-ranking of a minority of samples can cause a minor drop in S-Cor.






Compared with SBCA~\cite{li2026statistical}, SemanticFace achieves an 81.2\% reduction in MSE. Since SBCA is not open-sourced, its reported numbers are directly quoted from the original paper, and MMD and R-Precision are not included because they were not reported therein. Compared with DeadFace, our method yields an 89.3\% reduction in MSE, approximately a 50\% improvement in R-Precision, and an 83.5\% reduction in MMD, along with consistent gains across other evaluation metrics. These results highlight the benefit of aligning visual perception with language-grounded semantic representations in the interpretable ARKit space.

\begin{figure}[h]
    \centering
    \includegraphics[width=1.0\linewidth]{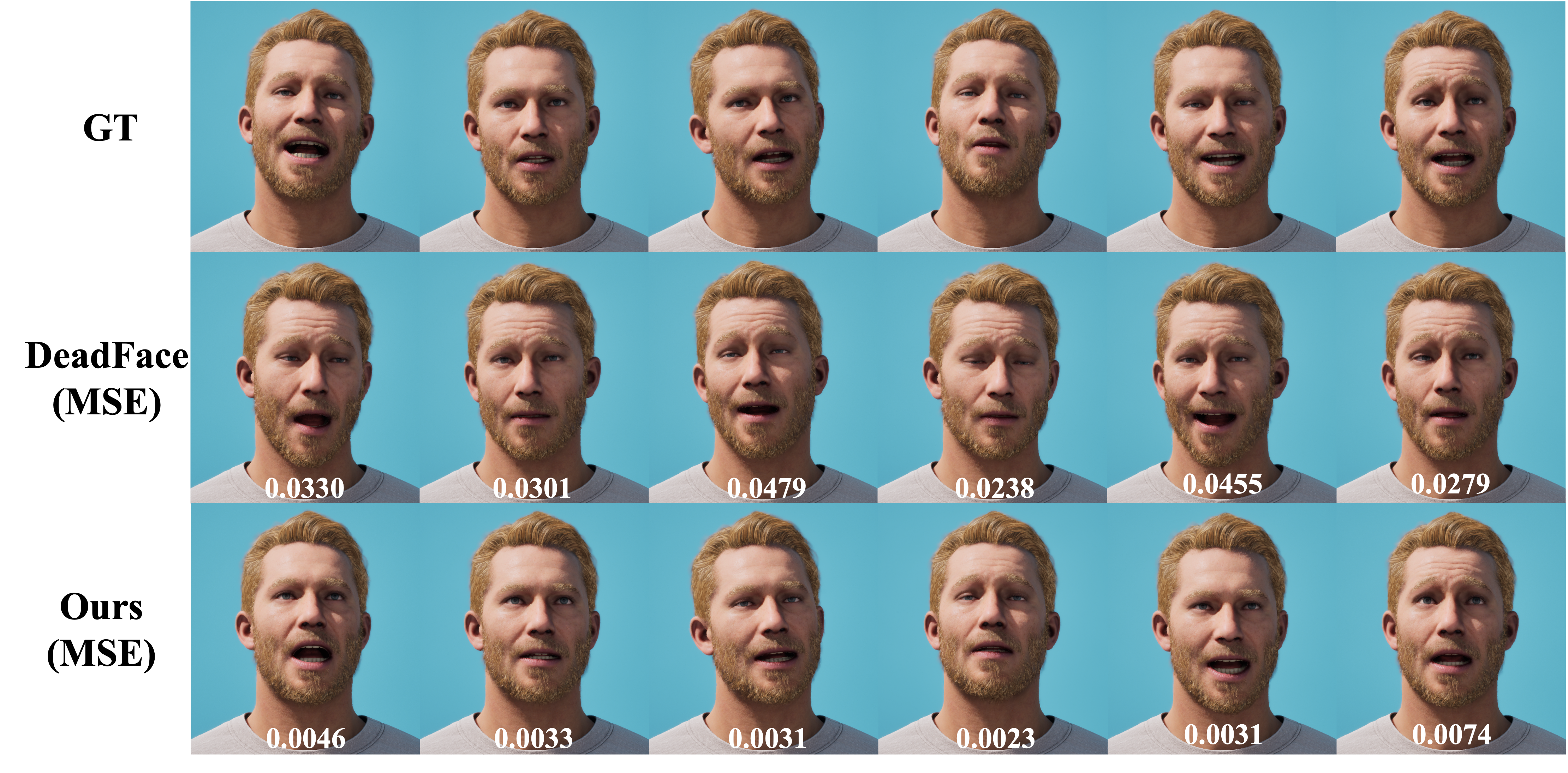} 
    \caption{
    \textbf{Qualitative Comparison on the Test Set with Ground-Truth ARKit coefficients.}
    From top to bottom: ground truth, DeadFace, and SemanticFace.
    Compared to the geometric baseline, our method produces facial actions that more closely match the ground-truth coefficients.
    The white numbers overlaid on each result denote the corresponding MSE values.
    }
    \label{fig:comparision_test}
\end{figure}

~\cref{fig:comparision_test} presents qualitative results across six representative facial expression cases. With the incorporation of semantic priors, our method produces more accurate and visually consistent facial actions, particularly in mouth closure, eye blinking, and eyebrow movements, consistently outperforming DeadFace in these regions. These observations indicate that language-aligned semantic supervision enhances both coefficient fidelity and perceptual realism, highlighting the advantage of reasoning in an interpretable action space.

\begin{table}[h]
\centering
\caption{Quantitative comparison with different methods on the in-the-wild data (R-Precision \& MMD). Best values are highlighted in \colorbox{red!20}{red}.}
\label{tab:wild_comparison}
\scriptsize
\setlength{\tabcolsep}{10pt}
\begin{tabular}{l ccc c}
\toprule
\multirow{2}{*}{\textbf{Method}}
& \multicolumn{3}{c}{\textbf{R-Precision} $\uparrow$}
& \multirow{2}{*}{\textbf{MMD} $\downarrow$} \\
\cmidrule(lr){2-4}
& \textbf{top-1} & \textbf{top-2} & \textbf{top-3} & \\
\midrule
DeadFace~\cite{deadface2023}
& 9.06\% & 19.90\% & 25.37\% & 0.90 \\
SemanticFace
& \cellcolor{red!20}23.38\% 
& \cellcolor{red!20}36.98\% 
& \cellcolor{red!20}47.10\% 
& \cellcolor{red!20}0.55 \\
\bottomrule
\end{tabular}
\end{table}

\subsection{Evaluation on In-the-Wild Data}
\label{sec:in-the-wild}




Unlike the test set, in-the-wild data does not provide 3D ground-truth coefficients. 
Consequently, metrics such as MSE and cross-comparison cannot be computed. 
We therefore evaluate performance using distribution-based metrics, namely R-Precision and MMD, which do not rely on explicit supervision.

Quantitative results are reported in \cref{tab:wild_comparison}. 
SemanticFace achieves higher R-Precision and lower MMD than DeadFace~\cite{deadface2023}, indicating improved alignment between predicted facial actions and the learned semantic distribution.

\begin{figure}[h]
    \centering
    \includegraphics[width=1\linewidth]{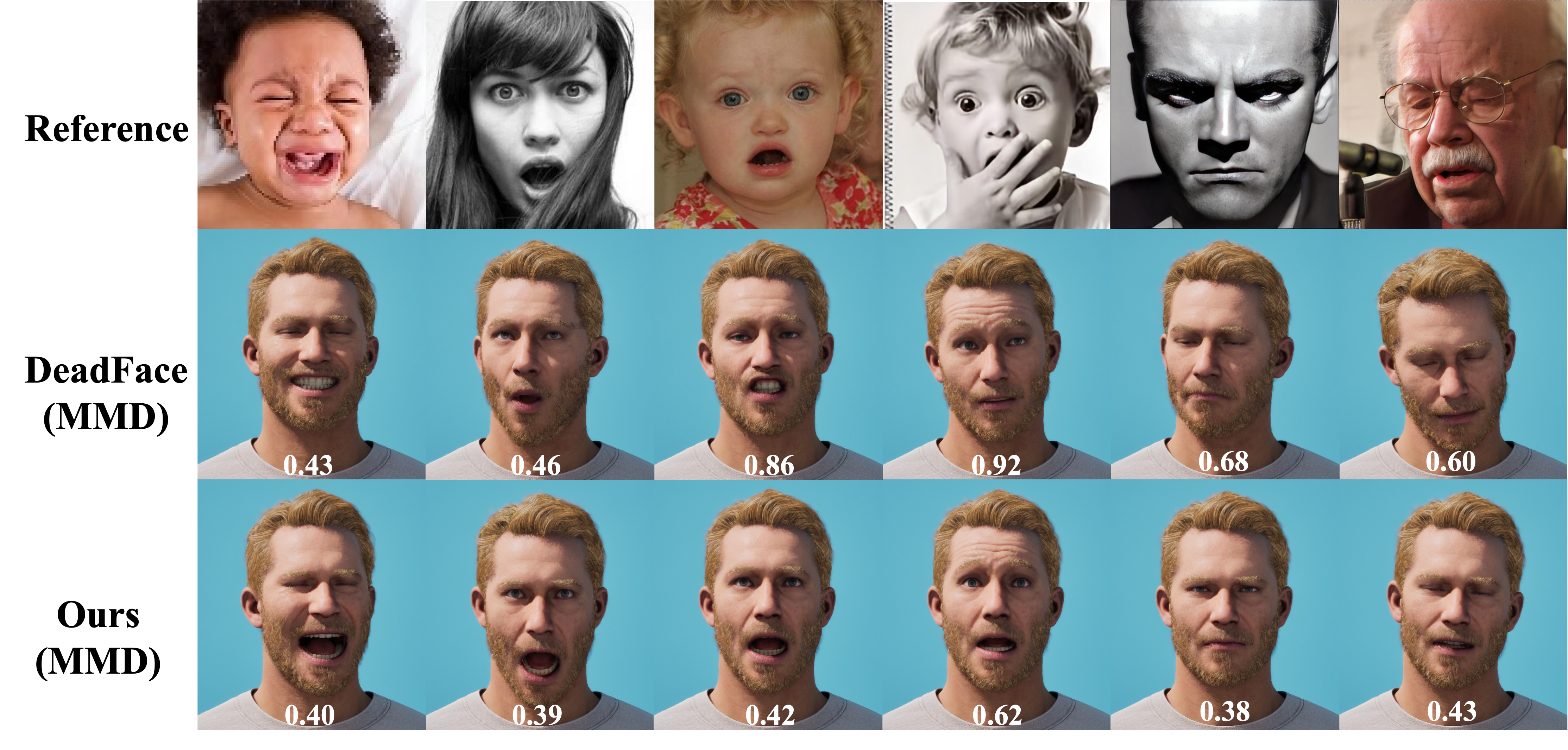}
    \caption{
        \textbf{In-the-wild qualitative comparison of interpretable facial action estimation.}
        SemanticFace maintains coherent and semantically consistent facial actions across diverse subjects and challenging conditions.
        The white numbers overlaid on each result denote the corresponding MMD values.
    }
    \label{fig:comparision_wild}
\end{figure}

Qualitative comparisons are shown in \cref{fig:comparision_wild}, covering six representative cases across diverse subjects (male, female, young, and elderly). 
By incorporating structured semantic priors, our method produces more coherent and perceptually consistent facial actions, particularly in mouth closure, eye blinking, and eyebrow movements. 
In contrast, DeadFace often exhibits temporal instability and local inconsistencies, partially attributable to upstream landmark noise from MediaPipe~\cite{lugaresi2019mediapipe}.

Furthermore, SemanticFace provides qualitative evidence of semantic consistency under challenging conditions such as occlusion by hair or hands and uneven illumination. 
Even without explicit 3D supervision, the model maintains semantically plausible and visually coherent predictions.


Overall, these results suggest that language-aligned semantic distillation improves distributional alignment and perceptual realism on in-the-wild dataset. 
Despite being trained on a relatively small number of subjects, the model demonstrates strong cross-identity generalization, producing semantically coherent facial actions across diverse in-the-wild images. 
As illustrated in \cref{fig:teaser}, the model first performs structured semantic reasoning before generating ARKit coefficients, enabling semantically grounded facial action prediction in unconstrained environments.

\subsection{Visual Comparison with FLAME-based Methods}
\label{sec:flame}

\begin{figure}[h] 
    \centering
    \includegraphics[width=0.95\linewidth]{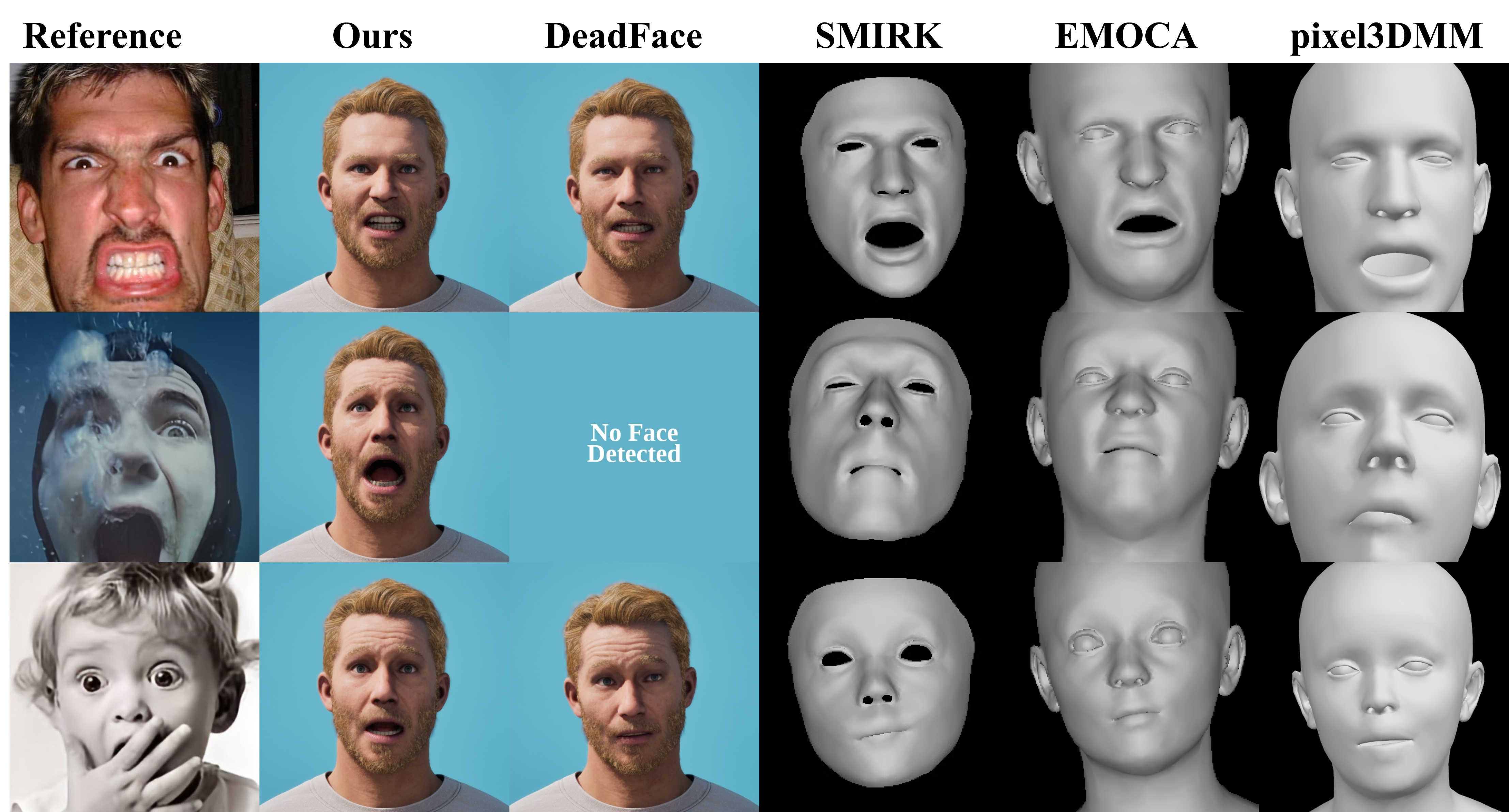} 
        \caption{\textbf{Comparison of ARKit-based and FLAME-based Methods on Hard Cases.} The input images are compared with results from SemanticFace (ours), DeadFace, SMIRK, EMOCA, and pixel3DMM.}
    \label{fig:flame_comparison}
\end{figure}

\begin{figure}[!h] 
    \centering
    \includegraphics[width=0.95\linewidth]{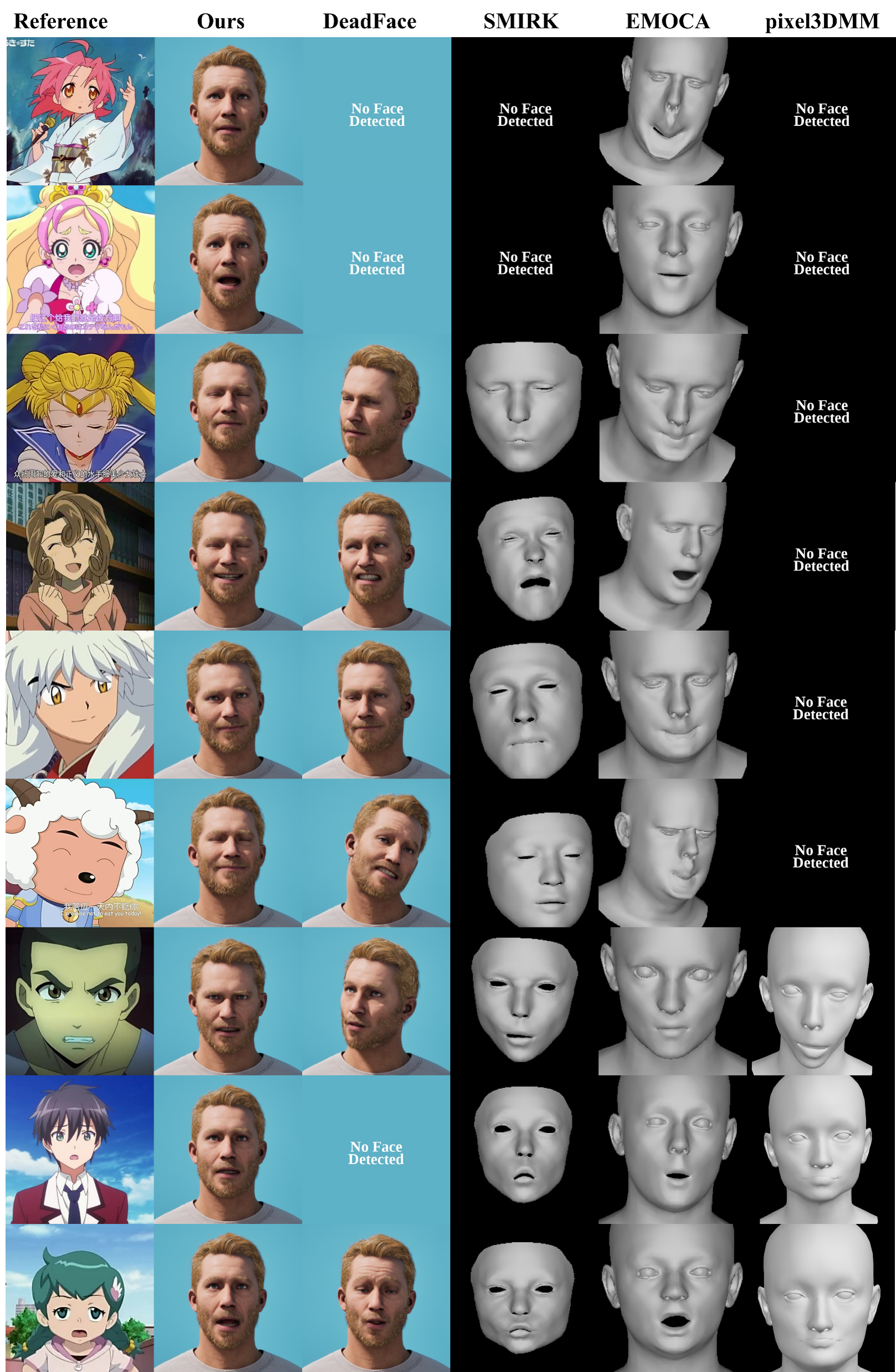} 
    \caption{\textbf{Qualitative comparison of larger domain shift.}}
    \label{fig:cartoon_comparison}
\end{figure}

Due to the absence of a reliable conversion between FLAME parameters and ARKit coefficients, we do not include quantitative comparisons with FLAME-based methods. Instead, we provide a qualitative comparison in ~\cref{fig:flame_comparison}. From left to right, the columns show the input image, SemanticFace (ours), DeadFace\cite{deadface2023}, SMIRK\cite{retsinas2024smirk}, EMOCA\cite{danvevcek2022emoca}, and pixel3DMM\cite{giebenhain2025pixel3dmm}. SemanticFace and DeadFace are ARKit-based methods, whereas SMIRK, EMOCA, and pixel3DMM are based on FLAME. The label ``No Face Detected'' indicates that the corresponding method fails to detect a face. FLAME-based methods can produce reasonable reconstructions on regular human faces, but their robustness degrades noticeably under extreme expressions, partial faces, and occlusions. In contrast, our method remains more stable in these challenging cases and better preserves the target facial expression.

We further evaluate robustness under a more extreme out-of-distribution setting, namely cartoon and stylized inputs, in Fig.~\ref{fig:cartoon_comparison}. We observe that the performance gap becomes even more pronounced under this larger domain shift. Our method remains capable of producing stable and semantically plausible facial actions on cartoon characters, whereas the other methods degrade substantially and often fail at the face detection stage, leading to the ``No Face Detected'' cases. These results further suggest that our ARKit-space semantic formulation provides stronger robustness not only on challenging real-face inputs, but also on non-photorealistic and stylized facial domains.

\subsection{Ablation Study}
\label{sec:ablation}



To quantify the contribution of semantic supervision, we evaluate three configurations with progressively increasing levels of semantic structure:




\begin{itemize}
    \item \textbf{A0: Coefficient Regression}.  
    The model predicts the ARKit coefficient vector 
    $\mathcal{V}=\{v_k\}_{k=1}^{K}$ directly from the input image, 
    without explicitly modeling the corresponding action identifiers 
    $\mathcal{A}=\{a_k\}_{k=1}^{K}$.

    \item \textbf{A1: Action–Value Prediction}.  
    The model predicts the structured action–value set 
    $\mathcal{S}=\{(a_k, v_k)\}_{k=1}^{K}$, 
    where action identifiers $a_k$ (e.g., \textit{EyeBlinkLeft}) 
    are provided as structured semantic tokens.

    \item \textbf{A2: Semantic Distillation (Ours)}.  
    In addition to the structured action–value set $\mathcal{S}$, 
    the model is supervised with hierarchical semantic descriptions 
    $T$ generated by the proposed two-stage semantic distillation framework.
\end{itemize}

These settings progressively introduce semantic structure from raw coefficients ($\mathcal{V}$) to action-value pairs ($\mathcal{S}$) and finally to hierarchical semantic descriptions ($T$).

\begin{table}[h]
\centering
\caption{Ablation analysis on semantic supervision evaluated on the test set.
All values are scaled by $10^{-3}$.
We report overall mean MSE, median MSE, standard deviation (Std), and 90th percentile (P90). Best values are highlighted in \colorbox{red!20}{red}.}
\label{tab:ablation_person2}
\scriptsize
\setlength{\tabcolsep}{4.5pt}
\begin{tabular}{l c c c c}
\toprule
\textbf{Method} 
& \textbf{Mean MSE} $\downarrow$
& \textbf{Median MSE} $\downarrow$
& \textbf{Std} $\downarrow$
& \textbf{P90} $\downarrow$ \\
\midrule
A0 
& 3.66 
& 3.44 
& 1.64 
& 5.74 \\

A1 
& 3.52
& 3.28
& 1.51 
& 5.57 \\

A2(Ours) 
& \cellcolor{red!20}3.34 
& \cellcolor{red!20}3.12 
& \cellcolor{red!20}1.49 
& \cellcolor{red!20}5.25 \\
\bottomrule
\end{tabular}
\end{table}

As reported in \cref{tab:ablation_person2}, performance improves monotonically from A0 to A2 across all statistics.
Both mean and median MSE decrease consistently, while the standard deviation and P90 are reduced, indicating improved robustness and fewer large-error cases.
Rather than relying solely on unconstrained numerical regression, semantic structure provides additional constraints that regularize the mapping from visual features to interpretable facial actions.



\begin{table}[h]
\centering
\caption{Paired t-test analysis on the test set using ablation settings A0, A1, and A2. Negative $\Delta$ indicates lower error for the first method in the comparison (i.e., $\Delta$ MSE is calculated as $\text{MSE}_{\text{first method}} - \text{MSE}_{\text{second method}}$). All improvements are statistically significant ($p<0.001$).}
\label{tab:ttest_person2}
\scriptsize
\setlength{\tabcolsep}{6pt}
\begin{tabular}{lccc}
\toprule
\textbf{Comparison} & \textbf{$\Delta$ MSE} $\downarrow$ & \textbf{t-statistic} & \textbf{p-value} \\
\midrule

A1 vs A0  
& $-1.42\times10^{-4}$ & $-5.28$ & $1.35\times10^{-7}$ \\

A2 vs A1  
& $-1.45\times10^{-4}$ & $8.21$ & $2.91\times10^{-16}$ \\

A2 vs A0  
& $-2.87\times10^{-4}$ & $10.61$ & $5.43\times10^{-26}$ \\

\bottomrule
\end{tabular}
\end{table}

To assess statistical significance, we conduct paired t-tests across test samples (\cref{tab:ttest_person2}).
All pairwise comparisons are statistically significant ($p<0.001$), providing evidence that the observed improvements are reliable and consistent across the dataset.

Overall, these results suggest that hierarchical semantic supervision improves both accuracy and stability by aligning visual representations with structured, interpretable facial action semantics.


\section{Conclusion}
\label{sec:conclusion}


We presented \textbf{SemanticFace}, a language-aligned framework for facial expression estimation in the interpretable ARKit blendshape space.
By distilling ARKit coefficients into structured semantic supervision, we reformulate expression estimation as semantic facial action prediction rather than purely numerical regression.
Extensive experiments demonstrate consistent improvements in coefficient accuracy, distributional alignment, and perceptual realism over existing baselines.
Notably, despite being trained on a limited number of subjects, the proposed framework demonstrates strong cross-identity generalization and maintains semantically coherent facial action prediction under diverse in-the-wild conditions.
More broadly, our findings suggest that language-aligned supervision provides a principled way to introduce semantic structure into interpretable facial representation learning.

\noindent\textbf{Limitations.}
SemanticFace operates within the predefined ARKit blendshape space, which may not fully capture subtle or highly non-linear facial deformations beyond the fixed basis. Furthermore, the structured semantic descriptions are generated by a pretrained language model and may inherit biases from its training data. Finally, Language-aligned supervision is more effective for interpretable facial muscle actions than for rigid head pose coefficients (e.g., \textit{HeadYaw}, \textit{HeadPitch}, \textit{HeadRoll}), as these geometric rotations lack clear linguistic abstraction and therefore benefit less from semantic distillation.


\clearpage
\bibliography{main}

\clearpage
\appendix

\section{Evaluation Metrics Detail}
\label{appendix:rprecision_mmd_training}

\subsection{Mean Squared Error (MSE)}
Let \(V_i = \{v_{i,k}\}_{k=1}^K\) denote the vector of ARKit blendshape coefficients for the \(i\)-th image \(\mathcal{I}_i\), where \(v_{i,k} \in \mathbb{R}\) is the activation level of the \(k\)-th facial action (\(K=61\)).  
Mean Squared Error (MSE) measures the average squared difference between the predicted coefficients \(V_i^{\text{pred}}\) and the ground-truth coefficients \(V_i^{\text{gt}}\). It is computed as

\[
\text{MSE} =
\frac{1}{N}\sum_{i=1}^{N}
\frac{1}{K}\sum_{k=1}^{K}
\left( v_{i,k}^{\text{pred}} - v_{i,k}^{\text{gt}} \right)^2.
\]

Lower values indicate higher coefficient prediction accuracy and better fidelity to the interpretable facial action space.

\subsection{R-Precision}
R-Precision is used to evaluate retrieval performance by examining whether the ground-truth motion corresponding to a given image description appears within the top-$K$(3) ranked candidates. For a batch of size $N$(32), we compute a cosine-similarity matrix $S_{ij}=\mathrm{sim}(\mathcal{I}_i, A_j)$ and determine the rank of the matched motion $A_i$ for each image query $\mathcal{I}_i$. R-Precision@K is then defined as
\[
\mathrm{R\text{-}Precision@K} = 
\frac{1}{N}\sum_{i=1}^{N}
\mathbf{1}[\mathrm{rank}(A_i | \mathcal{I}_i) \le K],
\]
indicating the proportion of queries whose correct matches fall within the top-$K$. Higher values reflect stronger cross-modal discriminative ability.

\subsection{Multimodal Distance (MMD)}
Multimodal Distance measures the geometric proximity between matched image and motion embeddings in the learned joint space. For each sample pair $(\mathcal{I}_i, A_i)$, we compute the Euclidean distance between their normalized embeddings and take the average over all pairs:
\[
\text{MMD} =
\frac{1}{N}\sum_{i=1}^{N}
\left\lVert f_{\text{image}}(\mathcal{I}_i) -
f_{\text{motion}}(A_i)
\right\rVert_2.
\]
Lower values indicate that the image and motion embeddings are located closer together, suggesting stronger semantic alignment.

\subsection{Evaluation Model Training Details}
The evaluation model is trained end-to-end using a symmetric InfoNCE loss, which encourages matched image–motion pairs to have high similarity while pushing unmatched samples apart. The loss is defined as
\[
\mathcal{L} = \tfrac{1}{2}(\mathcal{L}_{\mathcal{I}\rightarrow A} + 
\mathcal{L}_{A\rightarrow \mathcal{I}}),
\]
with a temperature parameter $\tau=0.07$, this is applied over all image–motion pairs within each batch. Training employs the AdamW optimizer with a learning rate of $1\times 10^{-4}$ and weight decay of the same magnitude, together with a cosine-annealing learning rate schedule. We train the model for up to 1000 epochs and apply early stopping when the R-Precision@1 on the validation split does not improve for ten consecutive epochs. Both the image encoder and motion encoder are fully trainable during optimization. Motion features are standardized using Z-score statistics computed from the training set, while image descriptions are tokenized with the Qwen3-VL-Embedding-8B \cite{qwen3vlembedding}. A fixed random seed (42) is used throughout to ensure reproducibility of the results.

The evaluation model is trained independently from SemanticFace and does not share parameters or supervision signals with the proposed framework. To avoid bias, evaluation training uses only ground-truth coefficients and image–motion pairs, without exposure to semantic descriptions.

\subsection{Cross-Comparison}
The evaluation matrix\cite{li2026statistical}, illustrated in the Cross-Comparison table, provides a comprehensive benchmark of the proposed statistical blendshape prediction model's performance against the ARKit baseline \cite{apple_arkit} \emph{only for the 13 most dominant and expressive blendshapes} (EyeBlinkLeft, EyeLookDownLeft, EyeLookInLeft, EyeBlinkRight, EyeLookDownRight, EyeLookInRight, JawOpen, Mouth-SmileLeft, MouthSmileRight, MouthUpperUpLeft, MouthUpperUpRight, BrowDownLeft, BrowDownRight). These 13 coefficients serve as primary indicators of facial expression quality. The table presents cross-comparison measures: P Corr \cite{pearson1901lines} denotes Pearson Correlation, quantifying the linear relationship between the outputs of the two models; S Corr \cite{spearman1910correlation} represents Spearman Correlation, evaluating the monotonic association; Accuracy indicates the overall prediction correctness based on thresholded expression detection; MSD refers to Mean Squared Deviation, capturing the average squared error from the ARKit reference. MSE measures the average on the total 61 coefficients while MSD foucses on 13 coefficients; and Deviation measures the mean absolute difference between the model predictions and the baseline.

\section{Prompt Design}
\label{appendix:prompt}

Our framework employs carefully designed prompts to enable semantic supervision generation and interpretable ARKit coefficient prediction. 
The prompts are designed to guide the large language model (LLM) and multimodal large language model (MLLM) to reason about facial muscle activations in a structured manner consistent with the Facial Action Coding System (FACS).

\subsection{Stage I: Semantic Supervision Generation}

In Stage I, the LLM converts ground-truth ARKit blendshape coefficients into structured semantic descriptions. 
The goal of this stage is to bridge the gap between low-level numerical parameters and high-level facial reasoning. 
Given ARKit coefficients, the model infers the corresponding facial expression category, detailed facial muscle movements, emotional implications when observable, and symmetry properties.

The prompt used in Stage I is shown below.

\begin{quote}
You are an expert in facial action coding (FACS).

Given ARKit blendshape coefficients, infer:

1. The most likely facial expression category.  
2. The detailed facial muscle movements.  
3. The emotional implication if clearly observable.  
4. The symmetry or asymmetry of the expression.

Base your reasoning only on facial muscle movement intensity.

Input: ARKit coefficient JSON.
\end{quote}

The input to this prompt is a JSON object containing all 61 ARKit coefficients, including eye movements, mouth movements, eyebrow actions, and head pose parameters. 
The LLM then generates a structured semantic description summarizing the facial action configuration implied by these coefficients. 
These descriptions serve as semantic supervision signals for Stage II training.

\subsection{Stage II: Language-Prior Semantic Distillation}

In Stage II, a multimodal large language model (MLLM) predicts ARKit coefficients directly from input facial images while being guided by the semantic reasoning paradigm introduced in Stage I. 
The prompt instructs the model to first analyze facial muscle activations in a structured semantic format and then generate the full set of ARKit coefficients.

The output consists of two parts:

\begin{itemize}
\item \textbf{analysis}: structured semantic reasoning about the facial expression.
\item \textbf{arkit}: a complete set of 61 ARKit blendshape coefficients.
\end{itemize}

To ensure deterministic and machine-readable outputs, the prompt enforces strict JSON formatting and requires that all coefficients be included.

The prompt used in Stage II is shown below.

\begin{quote}
You are an expert in Facial Action Coding System (FACS) and ARKit facial expression analysis.

Given a facial image, provide a two-part analysis.

Part 1: ``analysis''
Analyze the facial expression with the following structure:
\begin{itemize}
\item expression\_category: the most likely facial expression category
\item muscle\_movements: detailed facial muscle movements with intensity (by region: eyebrows, eyes, cheeks, mouth, jaw)
\item emotional\_implication: emotional meaning only if clearly observable from facial muscles
\item symmetry: describe left-right facial symmetry or any notable asymmetry
\end{itemize}
Part 2: ``arkit'' 
Generate a complete JSON object mapping every ARKit blendshape coefficient name to a precise value with three decimal places.

Requirements:
\begin{itemize}
\item Return only a valid JSON object with two keys: ``analysis''  and ``arkit'' .
\item The ``arkit''  field must contain all 61 ARKit coefficients.
\end{itemize}
\end{quote}

This structured prompting strategy enables the MLLM to align visual perception with interpretable facial action semantics while producing numerically precise ARKit parameters.

\section{Head Pose Analysis}
\label{appendix:head}

\begin{table}[h]
\centering
\caption{\textbf{Quantitative comparison of head pose parameters (HeadYaw, HeadPitch, HeadRoll) on the test set.} Best values are highlighted in \colorbox{red!20}{red}.}
\label{tab:head_comparison}
\scriptsize
\setlength{\tabcolsep}{3pt} 
\begin{tabular}{l cccc}
\toprule
\multirow{2}{*}{\textbf{Method}}
& \multicolumn{4}{c}{\textbf{Main Cross-Comparison}} \\
& \textbf{P-Cor} & \textbf{S-Cor} & \textbf{MSD} & \textbf{Deviation} \\
\midrule
HeadYaw(DeadFace\cite{deadface2023})
& 20.61\% & 21.03\% & 0.0074 & 0.0843 \\
HeadYaw(Ours) 
& \cellcolor{red!20}57.69\% & \cellcolor{red!20}55.60\% & \cellcolor{red!20}0.0048 & \cellcolor{red!20}0.0533 \\

\midrule
HeadPitch(DeadFace)
&19.60\%& 21.36\%&0.0350&0.0903\\
HeadPitch(Ours)
& \cellcolor{red!20}74.89\% & \cellcolor{red!20}72.09\% & \cellcolor{red!20}0.0045 & \cellcolor{red!20}0.0557 \\

\midrule
HeadRoll(DeadFace)
& \cellcolor{red!20}90.18\%
& \cellcolor{red!20}89.60\%
& \cellcolor{red!20}0.0009
& \cellcolor{red!20}0.0306 \\
HeadRoll(Ours)
& 53.40\% & 50.23\% & 0.0037 & 0.0599
\\
\bottomrule
\end{tabular}
\end{table}

To better understand the behavior of rigid head pose parameters, we compare DeadFace~\cite{deadface2023} and SemanticFace on the test set in \cref{tab:head_comparison}. The results show that language-aligned semantic supervision affects the three rotation dimensions differently.

For \textit{HeadYaw} and \textit{HeadPitch}, our method improves all reported metrics over the baseline. Pearson correlation increases from 20.61\% to 57.69\% for \textit{HeadYaw} and from 19.60\% to 74.89\% for \textit{HeadPitch}, while MSD and Deviation are both reduced. These results suggest that semantic distillation provides useful guidance for yaw and pitch prediction.

A likely reason is that yaw and pitch have relatively clear linguistic counterparts in natural descriptions, such as turning the head left or right, or looking upward or downward. Therefore, the semantic supervision generated in Stage~I can provide meaningful structural priors for these two dimensions.

In contrast, \textit{HeadRoll} shows a clear performance drop, with Pearson correlation decreasing from 90.18\% to 53.40\% and Spearman correlation decreasing from 89.60\% to 50.23\%. Compared with yaw and pitch, roll corresponds to lateral head tilt, which has much weaker semantic grounding in natural language and is rarely emphasized in either everyday facial descriptions or our training scripts.

As a result, semantic supervision provides limited useful guidance for roll prediction and may introduce additional ambiguity. This observation is consistent with the limitation discussed in the main paper: our framework is most effective for semantically interpretable facial actions, while rigid geometric pose parameters, especially \textit{HeadRoll}, benefit much less from language-aligned supervision.

\section{Fine-Grained Per-Coefficient Evaluation}

\begin{table}[htbp]
\centering
\caption{Quantitative evaluation of blendshape parameters (Part 1), sorted in ascending order of MSE.}
\label{tab:comparison_part1}
\begin{tabular}{lcccc}
\toprule
 & MSE & P Corr & S Corr & Deviation \\
\midrule
MouthRight & 0.0000 & - & - & 0.0000 \\
TongueOut & 0.0000 & - & - & 0.0000 \\
LeftEyeRoll & 0.0000 & - & - & 0.0000 \\
RightEyeRoll & 0.0000 & - & - & 0.0000 \\
JawRight & 0.0000 & -0.0003 & -0.0003 & 0.0025 \\
CheekPuff & 0.0000 & -0.0005 & -0.0005 & 0.0032 \\
MouthLeft & 0.0001 & -0.0031 & -0.0031 & 0.0083 \\
MouthFrownRight & 0.0002 & 0.2529 & 0.2367 & 0.0144 \\
MouthFrownLeft & 0.0002 & 0.2159 & 0.1968 & 0.0145 \\
EyeLookUpLeft & 0.0002 & 0.3537 & 0.3341 & 0.0155 \\
EyeLookUpRight & 0.0002 & 0.3537 & 0.3341 & 0.0155 \\
MouthShrugUpper & 0.0002 & 0.6421 & 0.4816 & 0.0155 \\
MouthRollLower & 0.0003 & 0.5860 & 0.5840 & 0.0167 \\
EyeLookOutLeft & 0.0004 & 0.6045 & 0.4329 & 0.0208 \\
MouthRollUpper & 0.0005 & 0.5105 & 0.4798 & 0.0227 \\
EyeSquintLeft & 0.0008 & 0.5283 & 0.4858 & 0.0276 \\
EyeSquintRight & 0.0008 & 0.5273 & 0.4849 & 0.0278 \\
MouthPressRight & 0.0008 & 0.3044 & 0.2299 & 0.0276 \\
MouthShrugLower & 0.0008 & 0.6138 & 0.4356 & 0.0281 \\
JawLeft & 0.0010 & 0.2012 & 0.1975 & 0.0300 \\
MouthPressLeft & 0.0010 & 0.2816 & 0.2200 & 0.0298 \\
EyeLookOutRight & 0.0010 & 0.4863 & 0.3927 & 0.0311 \\
EyeLookInLeft & 0.0014 & 0.5907 & 0.5123 & 0.0370 \\
MouthDimpleLeft & 0.0017 & 0.2240 & 0.2164 & 0.0385 \\
MouthClose & 0.0018 & 0.7754 & 0.7399 & 0.0429 \\
LeftEyeYaw & 0.0020 & 0.4663 & 0.4054 & 0.0424 \\
RightEyeYaw & 0.0020 & 0.4604 & 0.3985 & 0.0428 \\
JawForward & 0.0023 & 0.6601 & 0.6507 & 0.0454 \\
MouthDimpleRight & 0.0025 & 0.2056 & 0.1993 & 0.0451 \\
MouthFunnel & 0.0026 & 0.7588 & 0.7413 & 0.0494 \\
CheekSquintRight & 0.0029 & 0.7072 & 0.6977 & 0.0532 \\
\bottomrule
\end{tabular}
\end{table}

\begin{table}[htbp]
\centering
\caption{Quantitative evaluation of blendshape parameters (Part 2), sorted in ascending order of MSE.}
\label{tab:comparison_part2}
\begin{tabular}{lcccc}
\toprule
 & MSE & P Corr & S Corr & Deviation \\
\midrule
CheekSquintLeft & 0.0029 & 0.6985 & 0.6919 & 0.0531 \\
MouthUpperUpLeft & 0.0029 & 0.7378 & 0.6936 & 0.0488 \\
BrowDownRight & 0.0032 & 0.6638 & 0.5664 & 0.0544 \\
BrowDownLeft & 0.0032 & 0.6636 & 0.5664 & 0.0544 \\
EyeLookInRight & 0.0033 & 0.4145 & 0.3049 & 0.0534 \\
MouthUpperUpRight & 0.0035 & 0.7353 & 0.6903 & 0.0507 \\
MouthStretchLeft & 0.0036 & 0.6715 & 0.6399 & 0.0522 \\
JawOpen & 0.0037 & 0.8282 & 0.8020 & 0.0596 \\
HeadRoll & 0.0037 & 0.5340 & 0.5023 & 0.0599 \\
MouthStretchRight & 0.0038 & 0.6731 & 0.6427 & 0.0531 \\
NoseSneerLeft & 0.0040 & 0.5884 & 0.5546 & 0.0509 \\
HeadPitch & 0.0045 & 0.7489 & 0.7209 & 0.0557 \\
NoseSneerRight & 0.0046 & 0.6101 & 0.5772 & 0.0516 \\
MouthLowerDownRight & 0.0047 & 0.9410 & 0.9428 & 0.0674 \\
HeadYaw & 0.0048 & 0.5769 & 0.5560 & 0.0533 \\
MouthLowerDownLeft & 0.0051 & 0.9404 & 0.9423 & 0.0681 \\
MouthSmileLeft & 0.0054 & 0.8915 & 0.8351 & 0.0684 \\
BrowInnerUp & 0.0056 & 0.9431 & 0.8344 & 0.0747 \\
MouthPucker & 0.0059 & 0.8556 & 0.7165 & 0.0704 \\
MouthSmileRight & 0.0061 & 0.8823 & 0.8402 & 0.0728 \\
LeftEyePitch & 0.0064 & 0.7380 & 0.7181 & 0.0509 \\
RightEyePitch & 0.0064 & 0.7380 & 0.7181 & 0.0509 \\
EyeWideRight & 0.0066 & 0.6728 & 0.5264 & 0.0725 \\
EyeWideLeft & 0.0067 & 0.6725 & 0.5261 & 0.0725 \\
BrowOuterUpRight & 0.0080 & 0.9075 & 0.8480 & 0.0805 \\
BrowOuterUpLeft & 0.0080 & 0.9077 & 0.8489 & 0.0804 \\
EyeBlinkLeft & 0.0086 & 0.9165 & 0.7327 & 0.0703 \\
EyeBlinkRight & 0.0087 & 0.9162 & 0.7324 & 0.0705 \\
EyeLookDownLeft & 0.0172 & 0.8162 & 0.7961 & 0.0639 \\
EyeLookDownRight & 0.0172 & 0.8168 & 0.7965 & 0.0640 \\
\textbf{Average} & 0.0036 & 0.5966 & 0.5429 & 0.0456 \\
\bottomrule
\end{tabular}
\end{table}

\cref{tab:comparison_part1,tab:comparison_part2} report detailed results for all ARKit coefficients in terms of Mean Squared Error (MSE), Pearson correlation (P Corr), Spearman correlation (S Corr), and Deviation. These per-coefficient statistics provide a more fine-grained view of model behavior beyond the overall results in the main paper.

\textit{MouthRight}, \textit{TongueOut}, \textit{LeftEyeRoll}, and \textit{RightEyeRoll} are coefficients with undefined Pearson and Spearman correlations, denoted as ``-'' in the tables. This is because these coefficients are consistently zero in the dataset, and our model also predicts them as zero. Since correlation metrics require non-zero variance, they are undefined in such cases. However, the zero MSE and zero deviation values reflect perfect agreement between the predicted and ground truth values.

A similar but less extreme phenomenon appears for several low-activity coefficients, including \textit{JawRight}, \textit{CheekPuff}, and \textit{MouthLeft}. Although their MSE and Deviation are very small, their correlation scores remain close to zero. This mainly results from their limited variance in the dataset: these actions are rarely activated and stay near zero for most samples. Under such conditions, correlation metrics become unstable and are highly sensitive to small fluctuations, while error-based metrics better reflect the numerical closeness of the predictions.

Overall, the model performs better on prominent and frequently occurring facial actions, such as \textit{JawOpen}, \textit{MouthClose}, \textit{MouthFunnel}, \textit{MouthUpperUpLeft}, \textit{MouthUpperUpRight}, and \textit{CheekSquint}. By contrast, relatively lower performance is observed on subtler and less frequent coefficients, such as \textit{MouthFrownLeft}, \textit{MouthFrownRight}, \textit{MouthDimpleLeft}, \textit{MouthDimpleRight}, and \textit{JawLeft}. These results suggest that coefficients with clearer semantic meaning, larger activation range, and better training coverage are easier to predict reliably under our semantic distillation framework.



\bibliographystyle{splncs04}
\end{document}